\pdfoutput=1

\documentclass[11pt]{article}

\usepackage[table,xcdraw]{xcolor}
\usepackage[preprint]{acl}
\usepackage{times}
\usepackage{latexsym}

\usepackage[T1]{fontenc}

\usepackage[utf8]{inputenc}

\usepackage{microtype}
\usepackage{booktabs}

\usepackage{inconsolata}

\usepackage{graphicx}

\usepackage{booktabs}
\usepackage{adjustbox}
\usepackage{makecell}
\usepackage{algorithm}
\usepackage{algorithmic}
\usepackage{amsmath}
\usepackage{enumerate}
\usepackage[shortlabels]{enumitem}
\usepackage{xurl} 
\usepackage{hyperref}
\usepackage{listings}
\usepackage{float}
\usepackage{makecell} 
\usepackage{booktabs}
\usepackage{multirow}

\usepackage{listings}
\usepackage{xcolor}

\definecolor{codeboxcolor}{RGB}{240, 250,255}
\definecolor{codebordercolor}{RGB}{220, 239, 247}

\lstdefinestyle{prompt}{
    basicstyle=\ttfamily\small,  
    frame=single,                       
    backgroundcolor=\color{codeboxcolor},  
    rulecolor=\color{codebordercolor},  
    breaklines=true,                    
    breakatwhitespace=false,            
    breakindent=0pt,                    
    columns=fullflexible,               
    keepspaces=true,                    
    showstringspaces=false,             
    escapechar=|,                        
    lineskip=1.6pt, 
    linewidth=\linewidth,
    xleftmargin=0.01\linewidth,
    xrightmargin=0.01\linewidth,
}

\title{MathBuddy: A Multimodal System for Affective Math Tutoring}

\author{
  \textbf{Debanjana Kar\textsuperscript{*1,2}},
  \textbf{Leopold Böss\textsuperscript{*1}},
  \textbf{Dacia Braca\textsuperscript{*1}},
  \\
  \textbf{Sebastian Dennerlein\textsuperscript{1}},
  \textbf{Nina Hubig\textsuperscript{1}},
  \textbf{Philipp Wintersberger\textsuperscript{1}},
  \textbf{Yufang Hou\textsuperscript{1}}
\\
\\
  \textsuperscript{1}IT:U Interdisciplinary Transformation University Austria \\
  \textsuperscript{2}IBM Research India \\
  \small{
    \textbf{Correspondence:} 
    {debanjana.kar1@ibm.com, \{leopold.boess, dacia.braca\}@it-u.at}
  }
}

\begin{document}
\maketitle
\pagestyle{empty}
\thispagestyle{empty}
\pagenumbering{gobble}

\renewcommand{\thefootnote}{\fnsymbol{footnote}}
\footnotetext[1]{Authors contributed equally to this work.}
\renewcommand{\thefootnote}{\arabic{footnote}}

\begin{abstract}




The rapid adoption of LLM-based conversational systems is already transforming the landscape of educational technology. However, the current state-of-the-art learning models do not take into account the student's affective states. Multiple studies in educational psychology support the claim that positive or negative emotional states can impact a student's learning capabilities. To bridge this gap, we present \emph{MathBuddy}, an emotionally aware LLM-powered Math Tutor, which dynamically models the student's emotions and maps them to relevant pedagogical strategies, making the tutor-student conversation a more empathetic one. The student's emotions are captured from the conversational text as well as from their facial expressions. The student's emotions are aggregated from both modalities to confidently prompt our LLM Tutor for an emotionally-aware response. 
We have evaluated our model using automatic evaluation metrics across eight pedagogical dimensions 
and user studies. We report a massive 23 point performance gain using the win rate
and a 3 point gain at an overall level using DAMR scores which strongly supports our hypothesis of improving LLM-based tutor's pedagogical abilities by modeling students' emotions. Our dataset and code are available at: \url{https://github.com/ITU-NLP/MathBuddy}.\footnote{Link to demo video: \url{https://youtu.be/ZUjgmOw9GM0}}
\end{abstract}


\section{Motivation} \label{intro}

\begin{figure}[t]
    \centering    \includegraphics[width=\columnwidth]{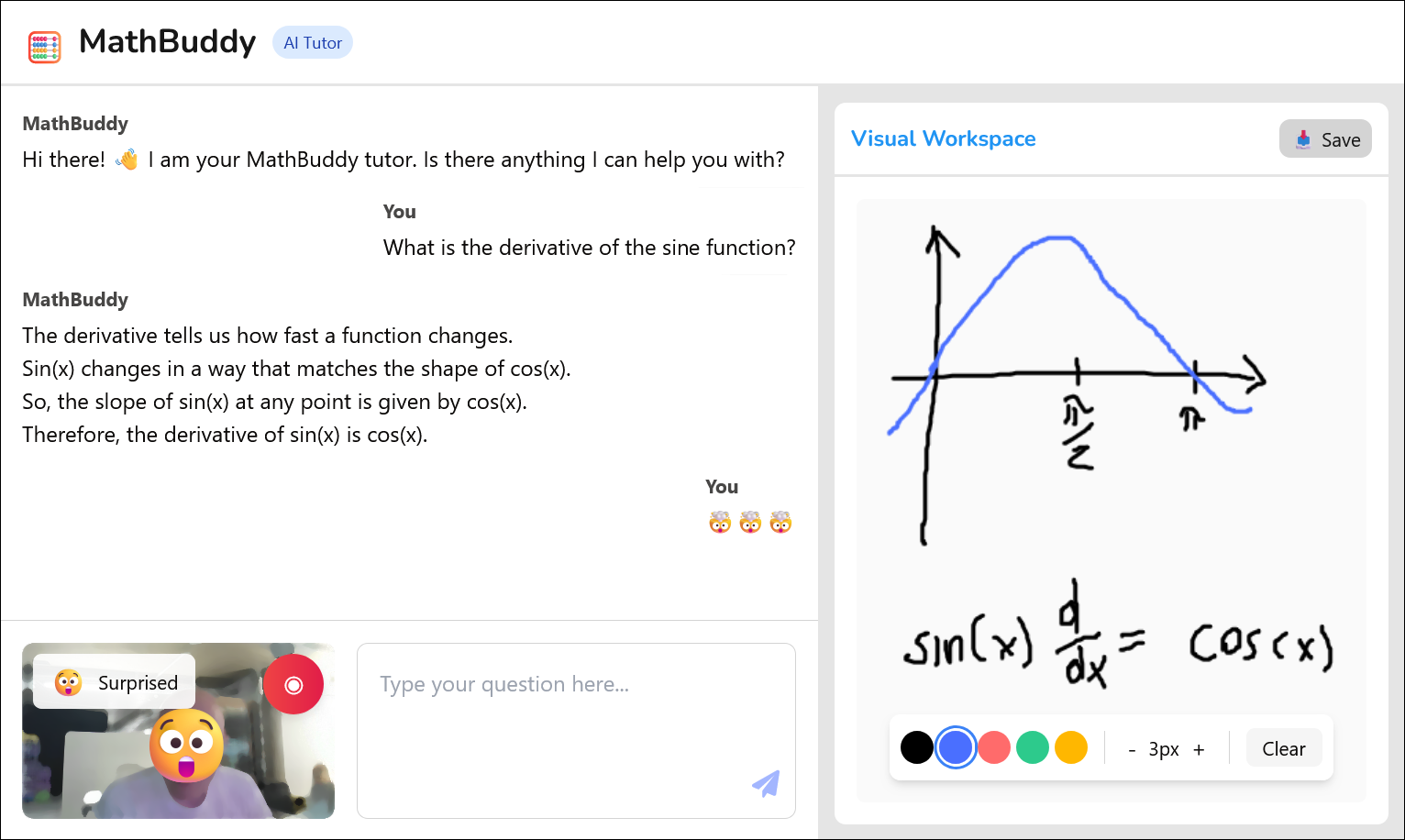}
    \caption{MathBuddy's user interface showcasing an example interaction using all functionality.}
    \label{fig:user_interface}
\end{figure}
The integration of AI tutors into educational platforms has rapidly evolved with the emergence of large language models (LLMs) \cite{chu2025llm}, enabling natural and interactive support for learners across subjects, such as mathematics \cite{macina2023mathdial}, 
and mitigating the issue of lack of qualified personal tutors.\footnote{\url{https://unesdoc.unesco.org/ark:/48223/pf0000385723}} Several studies \cite{10.1145/3657604.3662042, 10496545, WANG2024100247} claim that the integration of LLM-powered tutors facilitates formative self-regulated learning, where students learn and assess themselves at an independently decided pace.

Despite significant progress in instructional capabilities, most existing LLM-based tutoring systems lack emotional intelligence \cite{liu2024socraticlm, azerbayev2024llemma}, treating all learners homogeneously without accounting for their affective states. This can lead to disengagement, frustration, and ultimately suboptimal learning outcomes, especially in subjects such as mathematics, where learner anxiety is well-documented \cite{daker2023does, MAKI2024101316, schoenherr2025emotions}.

Research in educational psychology has long emphasized the role of emotions in learning \cite{pekrun2002academic, graesser2012intelligent}. Negative affective states, e.g., confusion and frustration, can inhibit problem-solving ability and motivation, while positive emotions, such as engagement and curiosity, are positively correlated with learning gains \cite{pekrun2006control, tan2021influence, zheng2023emotions}. 

At the same time, recent advances in NLP and computer vision have opened up new avenues for multimodal affect recognition. Transformer-based models can infer affective cues from text with increasing accuracy \cite{chutia2024review, ali2024transformers}, while facial expression recognition using deep learning has reached practical levels of robustness \cite{makhmudov2024enhancing}. 
However, existing LLM-based tutoring systems do not yet combine these modalities in real time to support accurate, empathetic adaptation in mathematics tutoring. 

To address this gap, we present our AI-based math tutor, \emph{MathBuddy} (see \autoref{fig:user_interface}), that models student emotions using both textual and visual modalities. 
More specifically, given a student's last utterance and facial expressions, \emph{MathBuddy} extracts the student's emotions from both modalities and aggregates them into one of three classes (\emph{Positive, Negative, or Neutral}). The emotion is then used to direct the LLM Tutor in one of the following ways: if the student is in a positive or neutral affective state, the student is challenged by \emph{MathBuddy}, while a negative affective state causes the system to try to motivate the student. 

We have evaluated the effectiveness of our system across eight different pedagogical dimensions \cite{maurya-etal-2025-unifying} through both automatic evaluation (Section \ref{sec:AutomaticEvaluation}) and real-time user studies (Section \ref{sec:UserStudy}). 
The results from both the evaluation strategies emphasize the importance of modeling student emotions in tutoring strategies.


Our work's main contributions are as follows:
\begin{enumerate}
    \item We present the first emotionally-aware LLM tutoring system, \emph{MathBuddy} that is grounded in education theory and adapts its response based on the student's affective state.
    \item We develop an automatic evaluation system to evaluate the framework thoroughly on a recent math tutor benchmark across eight different pedagogical dimensions based on the concepts introduced in \cite{maurya-etal-2025-unifying}.
    \item  We have annotated $224$ student utterances with emotion labels and polarity to support the development of emotion recognition models from text. 
\end{enumerate}


Our code, annotated dataset and a link to the system demo video are publicly available at: \url{https://github.com/ITU-NLP/MathBuddy}. 

\section{\emph{MathBuddy}: Design \& Implementation}
\label{sec:MathBuddyDesign}

In real-life tutoring scenarios, tutors often adapt their pedagogical strategies to the student's present emotional state \cite{lin2022good}.  
However, hardly any of the state-of-the-art AI-powered tutoring systems consider modeling students' emotions in this process.
Our proposed system, \emph{MathBuddy}, aims to address this gap by leveraging multimodal interaction channels. \emph{MathBuddy} models one-on-one interactions between a student and a tutor in the domain of Mathematics. In this section, we provide a detailed breakdown of the system’s implementation process.


\subsection{Emotion Recognition from Text and Webcam Data}

Since \emph{MathBuddy} is a conversational system, we try to gauge the student's affective state by extracting the emotion from the student's turn in the conversation. 
However, a major bottleneck we encountered in this task is the lack of such an annotated student-tutor conversational dataset. To overcome this challenge, we annotated the student turns in the hard version of the MathDial-Bridge dataset \cite{macina2025mathtutorbench}. 
Inspired by \citet{d2012language} work on affect in learning contexts, we assumed that students’ emotions can be multifaceted and vary in intensity. Hence, we modeled emotion extraction as a multi-label classification task with three target states: \emph{Boredom}, \emph{Engagement}, and \emph{Neutral} \citep{d2012language}.
To capture the intensity of each student’s emotional state at every turn, we assigned a polarity score on a scale from $0$ to $2$, where $0$ corresponds to low intensity and $2$ to high, thus serving as an indicator for arousal \cite{posner2005circumplex}.
Four annotators with academic backgrounds, aged between $27$ and $38$, participated in the annotation process. After an initial round, which yielded a Cohen’s Kappa inter-annotator agreement score of $40\%$, three annotators took part in the conflict adjudication process.
Finally, $224$ unique student turns were annotated ensuring an equal distribution across the labels.
We reserve the manually annotated dataset as our test set and generate a noisy student-tutor annotated dataset using DeepSeek-R1-Distill-LLama-70B \cite{deepseek2025deepseek} as our training dataset. We fine-tune multiple BERT-based models on the silver-labeled training data. 
The best model \cite{sanh2019distilbert} reports an accuracy and F1-score of $61.8$\% and $57.9$\% respectively. The detailed results are available in \autoref{sec:bert_evaluation}. 

In addition to text-based emotion recognition, the system processes webcam data to extract a student's affective states based on the student's facial expression. 

We use the \emph{face-api.js} package \cite{muehler2024faceapi} for the task of detecting students' emotions through their facial expressions. The \emph{face-api.js} package represents a JavaScript library that builds on \emph{tensorflow.js} to provide functionality related to human faces, such as face recognition, face landmark detection, and face expression recognition. 
The system utilizes its lightweight and fast in-browser face expression recognition.
It is configured to recognise the emotions \emph{Happy}, \emph{Sad}, \emph{Angry}, \emph{Surprised}, \emph{Fearful}, \emph{Disgusted}, and \emph{Neutral}, with \emph{Neutral} being the default when no face is being detected.
With our current configurations and considering the mapping explained later in \autoref{subsec:multimodal_emo_agg}, \emph{face-api.js} reports an accuracy of $76\%$ and an F1-score of $71\%$ on our test set. The test set comprises $151$ images from FERAC~\cite{rajasree2024ferac} and $38$ images from the Facial Emotion Recognition Dataset~\cite{training2023facial}, combined to ensure diversity across age groups as well as the inclusion of individuals diagnosed as non-neurotypical. 

Given that the system requires all emotional states to be attributed to the specific timestamp of a message, the face emotion samples have to be aggregated over the interval bounded by two messages.
For this step, the system considers only the changes in recognised emotion, grouping equal consecutive emotion states.
This way, each group can be assigned a duration as well as an age---the time from now to the last sample within the group.
Aggregation computes a weighted sum of the durations of all groups associated with one specific emotion, choosing the emotion with the highest sum as the result.
The required weights are derived from a half-life decay function employing a group's age and a fixed half-life of 120 seconds.
See \autoref{alg:emotion_agg} for a specific description.
\begin{algorithm}
\caption{Temporal emotion aggregation.} \label{alg:emotion_agg}
    \begin{algorithmic}[1]
        \REQUIRE Sequence of emotion samples $S = \{(e_1, t_1), \dots (e_n, t_n)\}$ sorted by time~$t$
        \STATE Initialize group list $G \leftarrow [\ ]$
        \STATE Current emotion $e_c \leftarrow e_1$, start time $t_s \leftarrow t_1$
        \FOR{$i = 2$ to $n$}
            \IF{$e_i \neq e_c$}
                \STATE Add new group $(e_c, t_s, t_{i-1})$ to $G$
                \STATE $e_c \leftarrow e_i$, $t_s \leftarrow t_i$
            \ENDIF
        \ENDFOR
        \STATE Add final group $(e_c, t_s, t_n)$

        \STATE Initialize emotion score map $S_e  \leftarrow \{\}$
        \STATE Half-life constant $\lambda \leftarrow {\ln(2)}/{120}$
        \FOR{each group $(e, t_s, t_e) \in G$}
            \STATE duration $d \leftarrow t_e - t_s$
            \STATE age $a \leftarrow \text{now} - t_e$
            \STATE weight $w \leftarrow \exp\left(-\lambda \times a \right)$
            \STATE $S_e[e] \leftarrow S_e[e] + d \times w$
        \ENDFOR

        \STATE $e^*  \leftarrow \arg\max_{e} S_e[e]$
        \RETURN $e^*$
    \end{algorithmic}
\end{algorithm}

\begin{figure*}[h]
    \centering
    \includegraphics[scale=0.78]{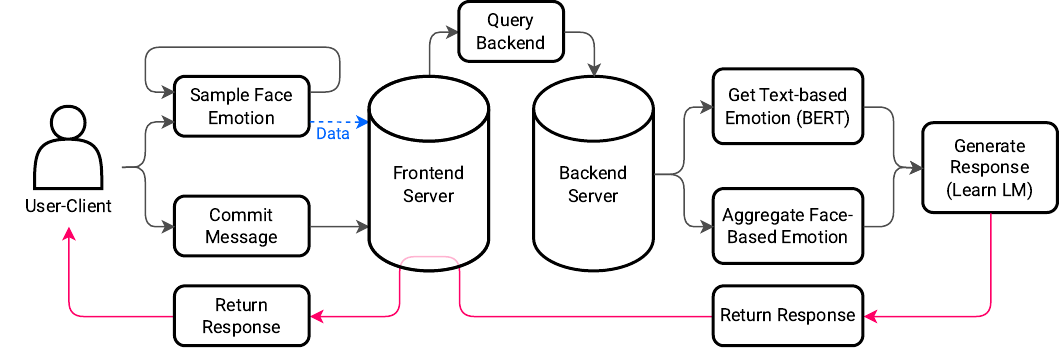}
    \caption{An overview of the system's logical flow, showcasing the core components.}
    \label{fig:system_flow}
\end{figure*}

\subsection{Multimodal Emotion Aggregation}\label{subsec:multimodal_emo_agg}

Users are assumed to feel only one unique emotion at any given time. Therefore, the system aims to approximate this emotion by compiling all sampled emotional states into a single one. 
Each modality ideally increases the accuracy of this aggregation.

A straightforward approach is to project the emotions sampled from both modalities to a common value space to allow them to be merged. 
This value space holds three primitive emotion states: \emph{Positive}, \emph{Neutral}, and \emph{Negative}. 

The emotion mapping applied by the system can be expressed as follows:
\begin{align*}
    m & : E \to P \\
    m(e) &= 
    \begin{cases}
        \text{Positive}, & \text{if } e \in E_{\text{positive}} \\
        \text{Neutral},  & \text{if } e \in E_{\text{neutral}} \\
        \text{Negative}, & \text{if } e \in E_{\text{negative}} 
    \end{cases}
\end{align*}

\noindent
where $E$ denotes the set of all recognizable emotions across modalities, and $P$ denotes the set of primitive categories: \emph{Positive}, \emph{Negative}, and \emph{Neutral}. Note that $P$ is a subset of $E$ used for simplified downstream processing. The specifics of how each emotion detected from either modality is mapped to one of the above primitive emotion states is detailed in \autoref{sec:emotion_mapping}. 
Each emotion state produced by either emotion recognition method can be directly mapped to these primitives, including their confidence values. Notably, states produced via face-based recognition are mapped before aggregation.
Merging the resulting values considers the following rule: \emph{Positive} and \emph{Negative} are always preferred over \emph{Neutral}; confidence is used as a secondary criterion if both values are not equal to \emph{Neutral}.
For example, a text-based positive emotion with a confidence of $50\%$ and a contradicting facial-based negative emotion with a confidence of $70\%$ are reconciled into a negative emotion.
Hence, the system assumes that any hint of non-neutral emotion implies that neutrality is not given.
This behavior appears beneficial as a non-neutral emotional state should enable better adjustment of the currently applied pedagogical strategy.

\subsection{Emotion-Aware LLM Tutor}

Inspired by educational psychology literature \cite{pekrun2002academic, graesser2012intelligent}, we design our system to map dynamically extracted student emotions to relevant pedagogical strategies in order to enhance the learning experience for the student. After extracting and aggregating multi-modal student emotions, we finally map it to relevant pedagogical strategies such that:
\begin{itemize}
    \item \emph{Positive} student emotion, the LLM tutor is prompted to \emph{challenge} the student;
    \item \emph{Neutral} or a \emph{Negative} emotion, the tutor is prompted to \emph{motivate} the student.
\end{itemize}

Please refer to \autoref{sec:prompts} for detailed prompts.




\subsection{System Implementation}

The system 
is divided into two core components: Frontend and Backend.
The Frontend represents a web server implemented in TypeScript that serves the web-based client and a public API, providing all functionality needed by the said client.
This way, all requests must go through the Frontend, while all other components remain hidden from the outside.
These requests mainly include the tutor response generation, when the user submits a message, and applying the face based emotion recognition to the webcam input.
The later can be configured to be handled in browser by \emph{face-api.js} or by the backend via our own model.
Additionally, the Frontend also represents the main storage, holding a primitive in-memory database.

The client employs a simple design consisting of two parts: a chat-based interface and a visual workspace.
The chat-based interface visualizes the tutor-student conversation similar to established LLM UIs.
In addition to a text field, this interface also features an optional webcam input.
The client uses this webcam input to query the user's emotional state in regular intervals; it can be configured to give live feedback on the recognised emotion.
The visual workspace offers space to take notes or sketch ideas via mouse input.
This feature is an artifact of the idea to submit handwritten notes to the tutor, reserved for future research, owing to implementation constraints.

The Backend component acts as a REST API, implementing all functionality related to inference via Python. 
It both runs local models and employs external APIs to respond to queries made by the Frontend.
This structure allows for adapting models or configurations at the Backend without necessitating changes to the Frontend.
Its key endpoints include the tutor response generation based on conversation history and face-based emotion states, as well as face-based emotion recognition based on an image. An overview of the system's logical flow is illustrated in Figure \ref{fig:system_flow}. 

\section{Automatic Evaluation} \label{sec:AutomaticEvaluation}

\begin{table*}[t]
\centering
\scriptsize
\caption{Comparison study using Win Rate and DAMR across eight different pedagogical dimensions. The plus (+) version of each baseline represents our models. Best reported metrics are highlighted in bold. The overall score is a combined mean of the DAMR scores across the eight dimensions. }
\label{tab:results}
\begin{adjustbox}{center}
\begin{tabular}{l|c|ccccccccc}
\toprule
\textbf{Model} & \makecell{\textbf{Win} \\ \textbf{Rate}} & \makecell{\textbf{Mistake} \\ \textbf{Identification}} & \makecell{\textbf{Mistake} \\ \textbf{Location}} & \makecell{\textbf{Answer} \\ \textbf{Disclosure}} & \makecell{\textbf{Providing} \\ \textbf{Guidance}} & \makecell{\textbf{Action-} \\ \textbf{ability}} & \makecell{\textbf{Human-} \\ \textbf{likeness}} & \textbf{Coherence} & \makecell{\textbf{Tutor} \\ \textbf{Tone}} & \makecell{\textbf{Overall} \\ \textbf{Score}} \\
\midrule
QSLM     & 0.30 & 0.71 & 0.92 & 0.41 & 0.22 & 0.88 & 0.56 & 0.58 & 0.86 & 0.64 \\
QSLM+    & \textbf{0.37} & 0.71 & 0.92 & 0.41 & 0.22 & 0.88 & 0.56 & 0.58 & 0.86 & 0.64 \\
\midrule
LlemmaMM  & 0.38 & 0.22 & 0.93 & \textbf{0.53} & \textbf{0.35} & 0.75 & 0.65 & 0.70 & 0.92 & 0.63 \\
LlemmaMM+ & \textbf{0.38} & \textbf{0.27} & \textbf{0.96} & 0.50 & 0.30 & \textbf{0.90} & \textbf{0.72} & \textbf{0.83} & \textbf{0.97} & \textbf{0.68} \\
\midrule
LearnLM  & 0.59 & 0.96 & 1.00 & 0.65 & 0.32 & 0.99 & 0.93 & 0.91 & 1.00 & 0.85 \\
LearnLM+ & \textbf{0.82} & \textbf{1.00} & \textbf{1.00} & \textbf{0.69} & \textbf{0.37} & \textbf{0.99} & \textbf{0.98} & \textbf{0.98} & \textbf{1.00} & \textbf{0.88} \\
\bottomrule
\end{tabular}
\end{adjustbox}
\end{table*}

\begin{table*}[h!]
\centering
\scriptsize
\caption{Comparison study using Win Rate and DAMR across eight different pedagogical dimensions with a more complex prompt.} 
\label{tab:appendix}
\begin{adjustbox}{center}
\begin{tabular}{l|c|ccccccccc}
\toprule
\textbf{Model} & \makecell{\textbf{Win} \\ \textbf{Rate}} & \makecell{\textbf{Mistake} \\ \textbf{Identification}} & \makecell{\textbf{Mistake} \\ \textbf{Location}} & \makecell{\textbf{Answer} \\ \textbf{Disclosure}} & \makecell{\textbf{Providing} \\ \textbf{Guidance}} & \makecell{\textbf{Action-} \\ \textbf{ability}} & \makecell{\textbf{Human-} \\ \textbf{likeness}} & \textbf{Coherence} & \makecell{\textbf{Tutor} \\ \textbf{Tone}} & \makecell{\textbf{Overall} \\ \textbf{Score}} \\
\midrule
QSLM     & 0.33                                                & 0.52                                                              & \textbf{0.94}                                               & 0.49                                                                  & 0.26                                                         & \textbf{0.86}                                             & 0.61                                                      & \textbf{0.68} & \textbf{0.92}                                         & 0.66                                                     \\
QSLM+    & \textbf{0.35}                                       & \textbf{0.57}                                                     & 0.93                                                        & \textbf{0.49}                                                         & \textbf{0.28}                                                & 0.84                                                      & \textbf{0.61}                                             & 0.67          & 0.91                                                  & \textbf{0.66}                                            \\ 
\midrule
LlemmaMM  & 0.41                                                & \textbf{0.86}                                                     & \textbf{1.00}                                               & \textbf{0.53}                                                         & \textbf{0.23}                                                & \textbf{0.99}                                             & 0.66                                                      & 0.62          & \textbf{1.00}                                         & \textbf{0.74}                                            \\
LlemmaMM+ & \textbf{0.42}                                       & \textbf{0.22}                                                     & \textbf{0.95}                                               & 0.49                                                                  & \textbf{0.32}                                                & \textbf{0.82}                                             & \textbf{0.66}                                             & \textbf{0.72} & \textbf{0.95}                                         & \textbf{0.64}                                            \\
\midrule
LearnLM  & 0.46                                                & 0.87                                                              & 0.99                                                        & 0.51                                                                  & 0.21                                                         & 1.00                                                      & 0.65                                                      & \textbf{0.63} & 1.00                                                  & 0.73                                                     \\
LearnLM+ & \textbf{0.46}                                       & \textbf{0.87}                                                     & \textbf{0.99}                                               & \textbf{0.57}                                                         & \textbf{0.24}                                                & \textbf{1.00}                                             & \textbf{0.67}                                             & \textbf{0.59} & \textbf{1.00}                                         & \textbf{0.74}                                            \\ 
\bottomrule
\end{tabular}
\end{adjustbox}
\end{table*}

For a comprehensive evaluation of the system, we evaluate \emph{MathBuddy} through automatic evaluation strategies and a user study. In this section, we detail the automatic evaluation strategies along with an analysis of the quantitative results. We evaluate \emph{MathBuddy} with different backend LLMs on an existing math tutor benchmark \cite{maurya-etal-2025-unifying} with emotion information from the input text only. We use the hard version of the MathDial Bridge dataset \cite{macina2023mathdial} containing $327$ human-annotated conversations using two different metrics:
\begin{itemize}
    \item \textbf{Win Rate}: Rate at which reward model \cite{macina2025mathtutorbench} prefers the LLM Tutor response over the ground truth response.
    \item \textbf{Desired Annotation Match Rate (DAMR)}: $\%$ of labels assigned to the tutor responses matching the desiderata described in \cite{maurya-etal-2025-unifying}.
\end{itemize}

Following \cite{maurya-etal-2025-unifying}, to calculate the DAMR scores, we have assigned labels \emph{"Yes", "No", "To some extent"} to the tutor responses using a round table of LLMs as Judges (LaaJ) (see \autoref{sec:eval_prompt} for the prompt details). A response gets a high DAMR score if the assigned label matches the desired label for the given pedagogical dimension 
as defined in \cite{maurya-etal-2025-unifying}.
The authors 
also report a poor correlation coefficient with two LLMs as judges across eight pedagogical dimensions. In our work, we have considered four different LLMs as judges, namely, Llama-3-3-70b-instruct \cite{grattafiori2024llama}, Deepseek-V3 \cite{liu2024deepseek}, Mixtral-8X22b-Instruct-v0.1 \cite{jiang2024mixtral}, Phi-4 \cite{abdin2024phi} and a fifth ensemble model through majority voting for the same pedagogical dimensions. To establish the reliability of the LaaJ models, we use the mentioned LLMs to generate the labels for each tutor utterance on the annotated dataset\footnote{\url{https://github.com/kaushal0494/UnifyingAITutorEvaluation/blob/main/MRBench/MRBench_V2.json}} \cite{maurya-etal-2025-unifying} and report the Spearman Correlation, Pearson Correlation and Accuracy for the same (Appendix \autoref{tab:laaj}). 
We find that the ensemble LaaJ model serves as the best evaluator with the highest correlations across four out of the eight dimensions and comparable results for the other dimensions. We use this model to report our DAMR scores in \autoref{tab:results}. We have used the current state-of-the-art tutoring models \cite{liu2024socraticlm, azerbayev2024llemma, learnlmteam2024learnlm} as baselines for comparison.

Through \autoref{tab:results},  
we can observe that our feature enhanced models perform consistently better compared to their respective baseline versions (for prompt used, check \autoref{sec:simple_prompt}). In the smaller models like Qwen2.5-7B-Socratic-LM \cite{liu2024socraticlm} and Llemma-7B \cite{azerbayev2024llemma}, we do observe some variations, especially with regard to the \emph{Mistake Identification} and \emph{Mistake Location} dimensions. However, with much larger models like LearnLM 2.0, we see that the feature enhanced version produces responses with more desirable pedagogical qualities in six out of the eight dimensions and is comparable for the rest of the dimensions. We found consistent results (\autoref{tab:appendix}) when we replicated this experiment with a more complex prompt (prompt is attached in \autoref{sec:complex_prompt}, intended to test the model's pedagogy instruction following skills). This indicates the importance of modeling student's emotions across all the pedagogical dimensions in the learning models. 
We occasionally observe variability in results when using smaller models — particularly with more complex prompts. This suggests that smaller models may struggle or become overwhelmed when presented with an excessive amount of information. 

\subsection{Ablation Studies} \label{sec:ablation}
\autoref{tab:ablation} highlights the importance of student emotion features mapped to education theory in enhancing the tutor's pedagogical capabilities. Across both the simple and complex prompts, prompting the model with relevant pedagogical strategies based on the detected emotion of the student results in a steep performance gain (by 9 points and 2 points through simple and complex prompts respectively). For the ablation study, we have used the Qwen2.5-7B-Socratic-LM \cite{liu2024socraticlm} model. Since it is a smaller model, providing only the emotional information misguides the model. 
However, when instructions are grounded in educational theory, for instance, directing that if a student exhibits boredom, the instructional response should aim to enhance the learner’s motivation, the model demonstrates a markedly stronger performance. 
\begin{table}[]
\centering
\small
\caption{Ablation study results with emotions and education theory features. Prompt 1 and Prompt 2 refer to the simple and complex prompts respectively. QSLM = Qwen2.5-7B-Socratic-LM; ET = Education Theory}
\label{tab:ablation}
\begin{tabular}{l|cc}
\toprule
Model               & Prompt1 & Prompt2 \\ \midrule
QSLM                & 0.28          & 0.33           \\
QSLM + emotion      & 0.21          & 0.12           \\
QSLM + emotion + ET & \textbf{0.37} & \textbf{0.35}  \\ \bottomrule
\end{tabular}
\end{table}

In general, across all dimensions, the emotion feature seems to have resulted in a performance gain particularly in \emph{humanlikeness}, \emph{actionability}, \emph{coherence} and \emph{tutor tone}. Each of these dimensions is related to a human's empathetic side, again indicating the significance of modeling student emotions in the models. Since we only have an annotated textual dataset, the quantitative results reported in \autoref{tab:results} only employed emotions extracted from textual conversation. 
We evaluate the multimodal system in real-time through a user study discussed in the following section (\autoref{sec:UserStudy}) using the best model (LearnLM+).

\section{User Study} \label{sec:UserStudy}

To evaluate the impact of emotion-aware adaptation on learning performance, we conducted a within-subject user study using \emph{MathBuddy}, our AI-based math tutoring system. 
The study aimed to assess whether integrating predicted student emotions leads to improved learning outcomes. A total of 30 participants (aged 15–55), representing diverse genders, nationalities, linguistic backgrounds, and educational profiles, took part in the study. Further details are provided in \autoref{sec: user_design_app}.

\subsection{User Study Analysis}
The collected user study comprises a diverse set of metadata and multimodal data: (1) learning outcomes, measured through multiple-choice questionnaires on select math topics (serving as a pre- and post-test); (2) interaction data, in the form of chat logs from both tutoring conditions (Emotion ON/OF modalities); (3) affective states, derived from emotion predictions based on textual and facial inputs; (4) learning experience, measured with a $15$-item post-interaction questionnaire including engagement and perceived support, rated on a $5$-point Likert scale; (5) overall satisfaction, captured through a final rating-scale survey; and (6) user-generated ground truth for the aggregated emotion, through participants’ gold annotations and corrections of the Emotion ON conversations.

\paragraph{Empathy is a desired quality in a tutor across pedagogical dimensions.}

\begin{figure}[h!]
    \centering
    \includegraphics[width=\linewidth]{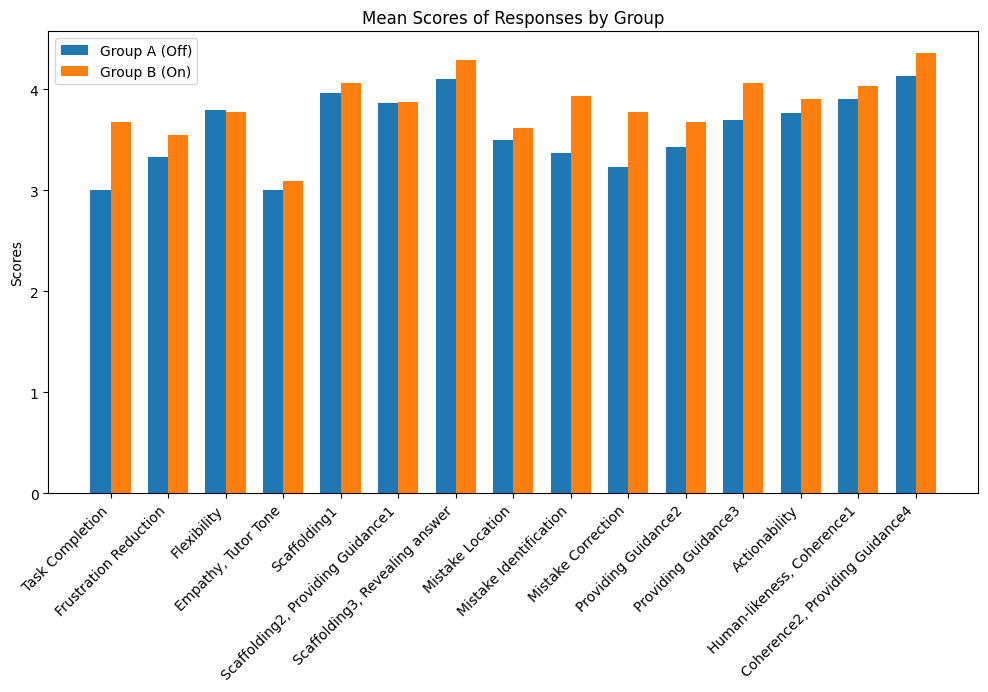}
    \caption{Average scores of participants with and without the emotion-aware condition}
    \label{fig:pedagogy}
\end{figure}

\autoref{fig:pedagogy} reports the mean score of $30$ participants across $15$ questions, each of them mapped to different pedagogical dimensions, the same discussed in \autoref{sec:AutomaticEvaluation} (\autoref{tab:mapqstn} in Appendix). 
Here, we observe that the average satisfaction scores are generally higher when emotions are used to guide \emph{MathBuddy}’s pedagogical strategies (Emotion ON condition). This suggests that participants reported a more positive perception of this interaction mode across the questionnaire dimensions, which we consider as indicators of their overall learning experience.

\paragraph{Empathetic responses enhances user learning.}



Comparing participants’ facial emotional dynamics over time, those in the Emotion ON condition displayed a higher frequency of positive expressions than those in the Emotion OFF condition (see \autoref{fig:facial_histogram} in the Appendix). 

\begin{figure}[h!]
    \centering
    \includegraphics[width=\linewidth]{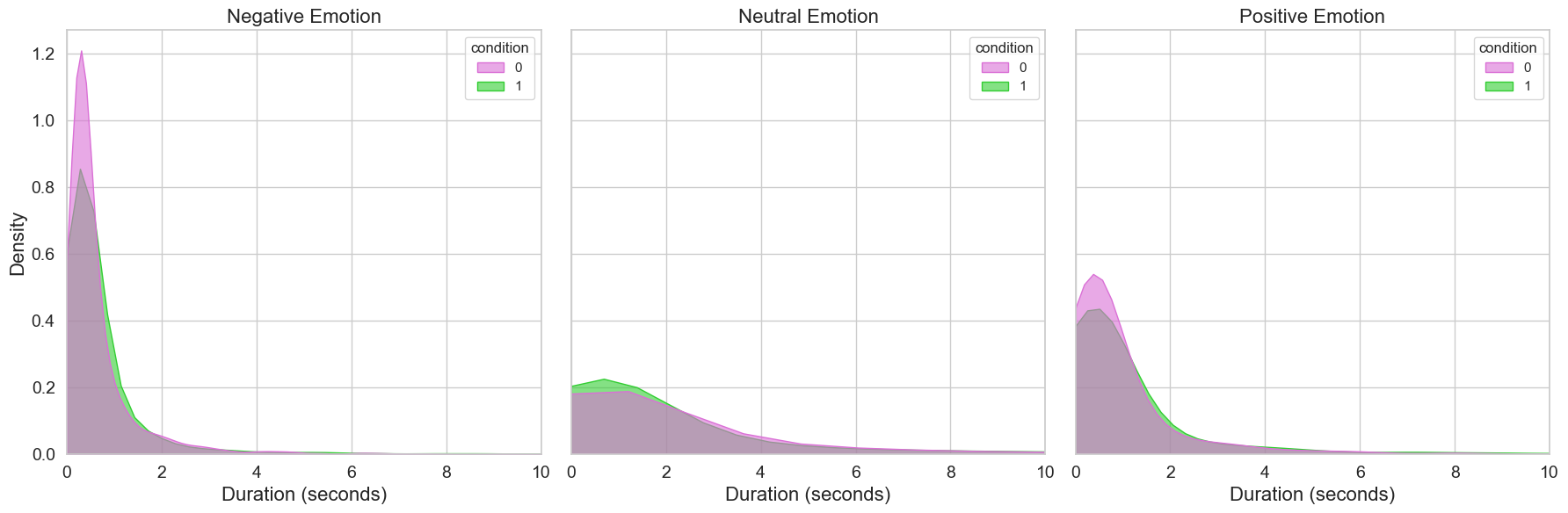}
    \caption{Kernel density estimates of facial emotion durations per class and condition.}
    \label{fig:kde_duration}
\end{figure}

Looking at the kernel density estimates (KDE) for the facial emotion duration distributions across conditions (\autoref{fig:kde_duration}), we can see that all three emotion classes (\emph{Negative}, \emph{Neutral}, \emph{Positive}) exhibit highly skewed distributions, with most durations clustered under 2 seconds. 

We calculated the mean duration per participant for each of the three facial emotional states and conducted Wilcoxon paired $t$-tests among the conditions; the results are summarized in \autoref{tab:duration_ttest}. We can appreciate that positive expressions showed a statistically significant difference ($p = .046$), with longer average durations under the Emotion ON condition ($1.14$s vs. $0.93$s).
This result indicates that participants experience longer positive emotional states under the Emotion ON condition.

These findings suggest that modeling student emotions and providing adaptive emotional feedback may have contributed to a more engaging and satisfying user experience during the tutoring sessions, potentially enhancing the overall learning process. 
Nonetheless, facial expressions commonly associated with happiness, such as smiling, can sometimes reflect more complex states, e.g., sarcasm or frustration. 
Expanding our multimodal channels could help capture emotional nuances more accurately, especially in technical domains such as mathematics, where affective signals are often subtle  and context-dependent.

\begin{table}[h]
    \centering
    \small
    \caption{Wilcoxon paired test on mean emotion durations per user and emotion (in seconds).}
    \label{tab:duration_ttest}
    \begin{tabular}{l|c|c|c}
        \toprule
        \textbf{Emotion} & \textbf{p-value} 
        & \makecell{\textbf{(Emo OFF)}\\\textbf{Mean $\pm$ Std}} 
        & \makecell{\textbf{(Emo ON)}\\\textbf{Mean $\pm$ Std}} \\
        \midrule
        Negative & 0.537          & 0.72 $\pm$ 0.56  & 0.65 $\pm$ 0.42 \\
        Neutral  & 0.248          & 12.73 $\pm$ 30.41 & 7.07 $\pm$ 14.7 \\
        Positive & \textbf{0.046} & 0.93 $\pm$ 0.87  & 1.14 $\pm$ 0.87 \\
        \bottomrule
    \end{tabular}
\end{table}



\paragraph{The detected emotions align with user emotions.}

As the last step of the experiment, the participants reviewed their own conversations in the Emotion ON condition, adjusting the tutor’s emotion labels when necessary. 
To assess the system's ability to correctly detect the participant’s emotions, we compare the aggregated text and facial emotions detected by the different modalities with the participant reviewed emotions. As reported in \autoref{tab:emotion_classification}, the system achieves an overall accuracy of $60$\% in real-time usage. However, we observe a sharp fall in the recall scores for the \emph{Neutral} class. This highlights the non-robust nature of our aggregation method where we try to suppress neutrality by preferring the modality that reports a non-neutral emotion. This can be improved by adopting a more sophisticated emotion aggregation method but remains to be explored as part of our future works.
\begin{table}[h!]
\centering
 \small
\caption{Comparison of system aggregated multimodal emotions to participant gold annotations.}
\label{tab:emotion_classification}
\begin{tabular}{l|ccc}
\toprule
\textbf{Emotion} & \textbf{Precision} & \textbf{Recall} & \textbf{F1-score} \\
\midrule
Negative & 0.57 & 0.98 & 0.72 \\
Neutral & 0.71 & 0.15 & 0.25 \\
Positive & 0.79 & 0.41 & 0.54 \\
\bottomrule
\end{tabular}
\end{table}

\section{Conclusion}
In this paper, we introduced \emph{MathBuddy}, an emotionally aware LLM-based math tutor that uses both text and facial expressions to model student emotions and deliver empathetic, pedagogically informed responses. By bridging affective cues with adaptive tutoring strategies, \emph{MathBuddy} enhances student engagement and learning effectiveness by a massive 23 points using win rate \cite{macina2025mathtutorbench} and 3 points at an overall level using DAMR scores as reported in \autoref{tab:results}. The usefulness of our approach is also supported through our carefully designed user studies as reported in \autoref{sec:UserStudy}.



Our evaluations highlight the benefits of integrating emotional awareness into LLM-driven education. We envision \emph{MathBuddy} as a step toward more human-centered, emotionally intelligent learning systems.


\section*{Limitations}


Generally, the development and evaluation of tutoring systems is limited by the lack of a golden standard or straightforward performance metrics. 
Qualitative analysis relies on feedback from users, who, however, may not be aware of a tutor system's ideal behavior---for example, instead of rating the educational prowess, they might rate user experience.
Moreover, the user study described in this paper captured an age group devoid of children, whose idea of ideal tutoring may be vastly different and feature concepts our evaluation misses entirely.
We also introduce a dataset with this work which currently contains coarse-grain emotion labels for student utterances. However, we feel that finer emotion labels capturing micro facial expressions may greatly benefit in capturing a much more nuanced student affective state. We leave this non-trivial task for future work.
Based on the data and user feedback, it appears that the modalities currently used (text-based and face-based) may not be informative enough to accurately capture a user's emotional state.  In technical education contexts, emotional expression in text is often subtle, unlike in other genres such as poetry \cite{hou-frank-2015-analyzing}.
This insight motivates future investigation into additional modalities, e.g., spoken audio, handwritten notes, or biometric sensors, which ideally considerably increase emotion recognition accuracy.
Further, spoken audio and handwritten notes more accurately represent the classical learning environment people might be more familiar with than a digital chat interface. We also want to address the aggregation strategy that we use for merging the emotion classes from different modalities in our system. The strategy can be further improved to handle neutral emotional states better. We leave this improvement for future work.


\section*{Acknowledgments}
We would like to acknowledge the contributions of Yesica Yanina Duarte who helped us design the user study and the interface of the system. We also want to acknowledge Prof. Daniel Klotz who helped us with his deep insights and interesting critiques on this project. We extend a hearty vote of thanks to all the enthusiastic participants of the user study who are mostly members of the IT:U Interdisciplinary Transformation University Austria.

\section*{Ethical Considerations}

This work acknowledges the ethical implications of our system's design and the data it collects. All conducted studies adhered to the GDPR \cite{goddard2017eu}, upholding all included principles, e.g., purpose limitation, data minimization, confidentiality, as well as lawful, fair, and transparent processing. 

Before data collection, informed consent was obtained, clearly outlining the purpose of the research, the voluntary nature of participation, and the right to withdraw at any time without consequence. All data were anonymized to protect participant identities. Special attention was given to exclusively collecting the information necessary to fulfill the functionality of real-time interaction. Although the system processes identifiable data such as image recordings of a participant's face, this is solely used for emotion recognition and never stored beyond model inference.
The entire process followed the institutional ethical guidelines set by the IT:U Interdisciplinary Transformation University Austria.


\bibliography{references}

\section*{Appendix}

\appendix

\section{Trained BERT Evaluation}\label{sec:bert_evaluation}

\autoref{tab:bert_model_perfomance} depicts the evaluation results of all text-based emotion recognition models investigated.

\begin{table}[h!]
\small
\centering
\caption{Text emotion classification model performance comparison based on accuracy and F1-score on silver-labeled test set.}
\label{tab:bert_model_perfomance}
\begin{tabular}{l|c|c}
\toprule
\textbf{Model} & \textbf{Accuracy} & \textbf{F1-Score} \\
\midrule
bert-base-uncased & 0.611 & 0.573 \\
distilbert-base-uncased & 0.618 & 0.579 \\
roberta-base & 0.576 & 0.538 \\
distilroberta-base & 0.617 & 0.579 \\
roberta-base-go\_emotions & 0.600 & 0.534 \\
\bottomrule
\end{tabular}
\end{table}

\section{Trained Face Recognition Models Evaluation}\label{sec:fer_evaluation}

We trained two models using a custom Convolutional Neural Network (CNN) architecture with 3 convolutional layers, one baseline and one with an attention mechanism. In addition, we evaluated three pre-trained Vision Transformer (ViT) models: \texttt{google-vit-base-patch16-224} models.\footnote{
\url{https://huggingface.co/JamesJayamuni/emotion_classification_v1.2}\\
\url{https://huggingface.co/jayanta/google-vit-base-patch16-224-cartoon-emotion-detection}\\
\url{https://huggingface.co/jayanta/google-vit-base-patch16-224-cartoon-face-recognition}
}

For fine-tuning, we considered two dataset configurations obtained by merging samples from three existing datasets. The first, \emph{small}, includes only data from FERAC ~\cite{rajasree2024ferac}, and FER ~\cite{training2023facial}. The second, \emph{large}, also incorporates additional samples from AffectNet YOLO dataset.  
We trained one version of each model on both dataset configurations and evaluated them on their corresponding test sets. Results shown in the Table \ref{tab:fer_classification_results}. 

\begin{table}[h!]
\small
\centering
\caption{Comparison of model performance on validation and test sets}
\label{tab:fer_classification_results}
\resizebox{\columnwidth}{!}{%
\begin{tabular}{l|cc|cc}
\toprule
\textbf{} & \multicolumn{2}{c|}{\textbf{Small Test Set}} & \multicolumn{2}{c}{\textbf{Large Test Set}} \\
\textbf{Model} & \textbf{Accuracy} & \textbf{F1-Score} & \textbf{Accuracy} & \textbf{F1-Score} \\
\midrule
CNN & 0.646 & 0.487 & 0.614 & 0.540 \\
CNN (+ Att) & 0.667 & 0.503 & 0.610 & 0.540 \\
ViT emo cls & 0.794 & 0.740 & 0.793 & 0.787 \\
ViT emo det & \textbf{0.809} & \textbf{0.756} & \textbf{0.797} & \textbf{0.793} \\
ViT face rec& 0.799 & 0.747 & 0.789 & 0.783 \\
\bottomrule
\end{tabular}%
}
\end{table}

\section{System Prompt Template}\label{sec:prompts}
The exact prompt template employed to elicit the tutor model's response is as follows:
\begin{lstlisting}[style=prompt]
|\textbf{Task:}| 
Below is an instruction that describes a task, paired with an input that provides further context. Write a response that appropriately completes the request.

|\textbf{Instruction:}|
1. You are an experienced math teacher and you are going to respond to a student in a useful and caring way.
2. Gently nudge the student towards the correct answer using guiding questions as your response.
3. Also consider the student's emotional state. 
- Positive emotions include engagement and joy.
- Neutral emotions include neutral and surprise.
- Negative emotions include angriness, boredom, confusion, contempt, disgust, fear, frustration, and sadness.
4. If the student's last response indicates:
- negative emotion, please motivate the student as a teacher.
- If the student's last response indicates positive emotion or neutral emotion, please challenge the student as a teacher.

|\textbf{Full Conversation:}|
{}

Sentiment based on Student's Facial Expression and Text Input (out of Positive, Neutral, Negative):
{}

|\textbf{Tutors Response:}|
{}
\end{lstlisting}







\section{Simple Prompt Template} \label{sec:simple_prompt}
This is an example of the simple prompt template that we tested our model with. This is an adapted version of \cite{macina2025mathtutorbench}'s scaffolding generation prompt.
\begin{lstlisting}[style=prompt]
|\textbf{Task:}| 
Below is an instruction that describes a task, paired with an input that provides further context. Write a response that appropriately completes the request.

|\textbf{Instruction:}|
1. You are an experienced math teacher and you are going to respond to a student in a useful and caring way. 
2. Gently nudge the student towards the correct answer using guiding questions as response. Also consider the student's emotional state. 
3. If the student's last response indicates:
- boredom, please motivate the student as a teacher
- engagement, please challenge the student as a teacher.

4. The student is trying to solve the following problem.
    
|\textbf{Full Conversation:}|
{}

|\textbf{Tutors Response:}|
{}
\end{lstlisting}


\section{Complex Prompt Template} \label{sec:complex_prompt}
This is an example of the complex prompt template that we also tested our model with. The complex prompt is more verbose containing specific pedagogical instructions to follow. This is an adapted version of \cite{macina2025mathtutorbench}'s pedagogy instruction following prompt. 
\begin{lstlisting}[style=prompt]
|\textbf{Task:}| 
Below is an instruction that describes a task, paired with an input that provides further context. Write a response that appropriately completes the request.

|\textbf{Instruction:}|
1. Be a friendly, supportive tutor. 
2.Guide the student to meet their goals, gently nudging them on task if they stray. 
3. Ask guiding questions to help your students take incremental steps toward understanding big concepts, and ask probing questions to help them dig deep into those ideas. 
4. Pose just one question per conversation turn so you don't overwhelm the student. 
5. Also consider the student's emotional state. If the student's last response indicates boredom, please motivate the student as a teacher. 
6. If the student's last response indicates engagement, please challenge the student as a teacher. 
7. Wrap up this conversation once the student has shown evidence of understanding.

|\textbf{Full Conversation:}|
{}

|\textbf{Tutors Response:}|
{}
\end{lstlisting}

\section{Evaluation Prompt Template} \label{sec:eval_prompt}
The exact prompt template is employed for the LLM as Judge evaluation is as follows:
\begin{lstlisting}[style=prompt]
|\textbf{Instructions:}| 
1. Following is the solution to the given math problem, the conversation history between the student and the tutor as the student tries to solve the given math problem, and the tutor's response to the student's last utterance. 
2. Evaluate the tutor's response on the defined paradigms. 
3. Also provide your reasoning for the evaluation.

|\textbf{Math Solution:}| 
{}

|\textbf{Conversation History:}| 
{}

|\textbf{Tutor Response:}| 
{}

|\textbf{Response}| (Stick to the given format, output it as a json only): 

"Mistake identification": "Yes/No",
"Mistake location": "Yes/No", 
"Revealing of the answer": "Yes/No",
"Providing guidance": "Yes/No"
"Actionability": "Yes/No"
"Coherence": "Yes/No",
"Tutor tone": "encouraging/neutral/offensive"
"Human-likeness": "Yes/No"
"Reasoning": "The tutor's response is evaluated as follows..."

\end{lstlisting}

\section{Emotion Mapping} \label{sec:emotion_mapping}
The emotion mapping applied by the system can be expressed as follows:
\begin{align*}
    m & : E \to P \\
    E_{\text{positive}} &= \{ \text{Engaged, } \text{Happy} \} \\
    E_{\text{neutral}}  &= \{ \text{Neutral, } \text{Surprised} \} \\
    E_{\text{negative}} &= \{ \text{Angry, } \text{Boredom, } \text{Disgusted, }\\
    &\qquad \text{Fearful, } \text{Sad} \} \\
    m(e) &= 
    \begin{cases}
        \text{Positive, } & \text{if } e \in E_{\text{positive}} \\
        \text{Neutral, }  & \text{if } e \in E_{\text{neutral}} \\
        \text{Negative, } & \text{if } e \in E_{\text{negative}}
    \end{cases}
\end{align*}

\noindent
where $E$ denotes the set of all recognizable emotions across modalities, and $P$ denotes the set of primitive categories: \emph{Positive}, \emph{Negative}, and \emph{Neutral}. We labeled \emph{surprise} as neutral due to dataset limitations: since it can indicate both positive and negative emotions and no contextual information was available, this choice ensured consistency in the educational strategy. $E_{\text{positive}}$ and $E_{\text{negative}}$ denote the the positive and negative emotions detected through both modalities respectively. The emotions detected through text include \emph{Engagement, Neutral and Boredom} while facial expression emotions include \emph{Happy, Sad, Surprised, Angry, Disgusted and Fearful}. Note that $P$ is a subset of $E$ used for simplified downstream processing.

\section{LLM as a Judge Evaluation} \label{laaj}
Please refer to \autoref{tab:laaj}.
\begin{table*}[]
\small
\centering
\caption{LLM as Judge results}
\label{tab:laaj}
\begin{tabular}{c|ccccc}
\toprule
\textbf{Dimensions}                           & \textbf{Llama3-70B} & \textbf{DeepseekV3} & \textbf{Mixtral-22B} & \textbf{Phi4} & \textbf{Ensemble} \\ \midrule
\textbf{Tutor\_Tone\_spearman}                & 0.2965              & \textbf{0.5791}     & 0.2484               & 0.1808        & 0.4779            \\
\textbf{Tutor\_Tone\_pearson}                 & 0.297               & \textbf{0.5789}     & 0.2485               & 0.1819        & 0.4779            \\
\textbf{Tutor\_Tone\_accuracy}                & 0.5019              & \textbf{0.7907}     & 0.6522               & 0.4006        & 0.7025            \\ \midrule
\textbf{Humanlikeness\_spearman}              & \textbf{0.3605}     & 0.2377              & 0.0717               & 0.3548        & 0.3506            \\
\textbf{Humanlikeness\_pearson}               & 0.4101              & 0.2703              & 0.0757               & 0.4494        & \textbf{0.4264}   \\
\textbf{Humanlikeness\_accuracy}              & 0.8758              & 0.8832              & 0.582                & 0.8857        & \textbf{0.887}    \\ \midrule
\textbf{Mistake\_Identification\_spearman}    & \textbf{0.6218}     & 0.508               & 0.1351               & 0.3317        & 0.5914            \\
\textbf{Mistake\_Identification\_pearson}     & \textbf{0.6442}     & 0.5311              & 0.1396               & 0.3579        & 0.6147            \\
\textbf{Mistake\_Identification\_accuracy}    & \textbf{0.8571}     & 0.8193              & 0.5615               & 0.8112        & 0.8516            \\ \midrule
\textbf{Mistake\_Location\_spearman}          & 0.3199              & 0.3756              & 0.169                & 0.1981        & \textbf{0.4019}   \\
\textbf{Mistake\_Location\_pearson}           & 0.319               & 0.3997              & 0.1704               & 0.2169        & \textbf{0.4268}   \\
\textbf{Mistake\_Location\_accuracy}          & 0.5155              & 0.5988              & 0.5205               & 0.6373        & \textbf{0.628}    \\ \midrule
\textbf{Revealing\_of\_the\_Answer\_spearman} & 0.8195              & 0.7925              & 0.5543               & 0.7854        & \textbf{0.821}    \\
\textbf{Revealing\_of\_the\_Answer\_pearson}  & 0.8195              & 0.794               & 0.5582               & 0.7891        & \textbf{0.8206}   \\
\textbf{Revealing\_of\_the\_Answer\_accuracy} & 0.9379              & 0.9373              & 0.8957               & 0.9385        & \textbf{0.9516}   \\ \midrule
\textbf{Providing\_Guidance\_spearman}        & 0.3824              & 0.3674              & 0.1575               & 0.3923        & \textbf{0.4302}   \\
\textbf{Providing\_Guidance\_pearson}         & 0.4037              & 0.4278              & 0.1689               & 0.4384        & \textbf{0.4858}   \\
\textbf{Providing\_Guidance\_accuracy}        & 0.6031              & 0.5807              & 0.4547               & 0.6199        & \textbf{0.6217}   \\ \midrule
\textbf{Actionability\_spearman}              & 0.3122              & 0.307               & 0.2058               & 0.382         & \textbf{0.3937}   \\
\textbf{Actionability\_pearson}               & 0.3081              & 0.3477              & 0.2097               & 0.4081        & \textbf{0.4162}   \\
\textbf{Actionability\_accuracy}              & 0.6224              & 0.5894              & 0.5366               & 0.6354        & \textbf{0.6503}   \\ \midrule
\textbf{Coherence\_spearman}                  & \textbf{0.4696}     & 0.392               & 0.1058               & 0.3975        & 0.4165            \\
\textbf{Coherence\_pearson}                   & \textbf{0.5102}     & 0.4591              & 0.1215               & 0.4744        & 0.4739            \\
\textbf{Coherence\_accuracy}                  & \textbf{0.8453}     & 0.8354              & 0.5652               & 0.8354        & 0.8398            \\ \bottomrule
\end{tabular}
\end{table*}

\section{User Study Questions} \label{user_study_q}
Please refer to \autoref{tab:mapqstn}.

\begin{table*}[h]
\centering
\caption{Mapping of User Study Questions to relevant Pedagogy Dimensions and metrics.}
\label{tab:mapqstn}
\begin{tabular}{l|c}
\toprule
\textbf{Question}                                                  & \textbf{\begin{tabular}[c]{@{}c@{}}Pedagogy Dimension\\ /Metric\end{tabular}} \\ \midrule
I completed the math problem on time.                              & Task Completion                                                               \\ \midrule
The tutor helped me feel less stuck or frustrated.                 & \begin{tabular}[c]{@{}c@{}}Frustration \\ Reduction\end{tabular}              \\ \midrule
I felt the tutor adapted to my needs.                              & Flexibility                                                                   \\ \midrule
The tutor seemed aware of how I was feeling.                       & \begin{tabular}[c]{@{}c@{}}Empathy, \\ Tutor Tone\end{tabular}                \\ \midrule
The tutor guide me toward the solution.                            & Scaffolding1                                                                  \\ \midrule
The tutor helped me think through the problem step by step.        & \begin{tabular}[c]{@{}c@{}}Scaffolding2, \\ Providing Guidance1\end{tabular}  \\ \midrule
The tutor guided me without simply giving away the answer.         & \begin{tabular}[c]{@{}c@{}}Scaffolding3, \\ Revealing answer\end{tabular}     \\ \midrule
The tutor helped me understand where I made mistakes.              & Mistake Location                                                              \\ \midrule
The tutor helped me identify what my mistake was.                  & Mistake Identification                                                        \\ \midrule
The tutor helped me to correct my mistakes.                        & Mistake Correction                                                            \\ \midrule
The tutor helped me understand math concepts better.               & Providing Guidance2                                                           \\ \midrule
I was able to connect the explanations to what I already knew.     & Providing Guidance3                                                           \\ \midrule
The tutor’s feedback was useful and helped me know wat to do next. & Actionability                                                                 \\ \midrule
The tutor’s responses were coherent with what I asked.             & \begin{tabular}[c]{@{}c@{}}Human-likeness, \\ Coherence\end{tabular}          \\ \midrule
The explanations were easy to follow.                              & \begin{tabular}[c]{@{}c@{}}Coherence, \\ Providing Guidance4\end{tabular}     \\ 
\bottomrule
\end{tabular}
\end{table*}

\section{User Study Design} \label{sec: user_design_app}
A total of $30$ participants (aged 15–55), representing diverse genders, nationalities, linguistic backgrounds, and educational profiles, took part in the study. Each participant was assigned a unique user ID to ensure anonymity. 
Demographic distribution is reported in \autoref{fig:participants}.

Each participant completed two tutoring sessions with \emph{MathBuddy}, solving one geometry problem and one probability problem. In both sessions, the system captured facial expressions via webcam and analyzed textual responses to infer the student’s emotional state. However, only in the \textbf{Emotion ON} condition were these emotional predictions actively used to guide the tutor’s communication and instructional strategy. In the \textbf{Emotion OFF} condition, emotion detection was passive and did not influence the tutor’s behavior. 
The two math problems and the order of conditions were randomized across participants to control for order and content effects. Participants were aware of the existence of two different system configurations but were blind to their order.
Each session lasted a maximum of 10 minutes. Participants could use a digital whiteboard, chat freely with the AI tutor, and decide whether to explore the solution after solving the problem or end the session early. Before and after the two sessions, participants completed a 6-question multiple-choice test to assess baseline and post-interaction knowledge in geometry and probability. Each test was independently completed within a 5-minute time limit.

After each tutoring session, participants filled out a 15-item Likert-scale questionnaire evaluating their experience in terms of engagement, clarity, frustration, emotional alignment, and perceived helpfulness of the tutor. At the end of the experiment, they completed a final satisfaction survey (4 closed and 4 open-ended questions), where they could share general feedback about the system, their experience, and any perceived limitations.

Additionally, participants were shown their chat transcript from the Emotion ON session and asked to highlight tutor responses they found particularly helpful—either in terms of mathematical support or emotional alignment. They were also invited to review and correct the system’s predicted emotions, enabling us to gather feedback on the accuracy of emotion recognition.

The entire experiment lasted between 30 and 45 minutes. All participants were presented with the same math problems and knowledge test items, carefully selected based on a pre-study survey. In this preliminary phase, 25 respondents (aged 24–45, primarily from academic environments) identified which mathematical topics they had found most difficult during their education. While calculus-related topics (e.g., derivatives, integrals) were the most frequently cited, geometry and probability followed closely and were chosen for their conceptual richness and suitability for short, interactive sessions.




\section{Participant Details} \label{participant}
\begin{figure*}[h!]
    \centering
    \includegraphics[scale=0.23]{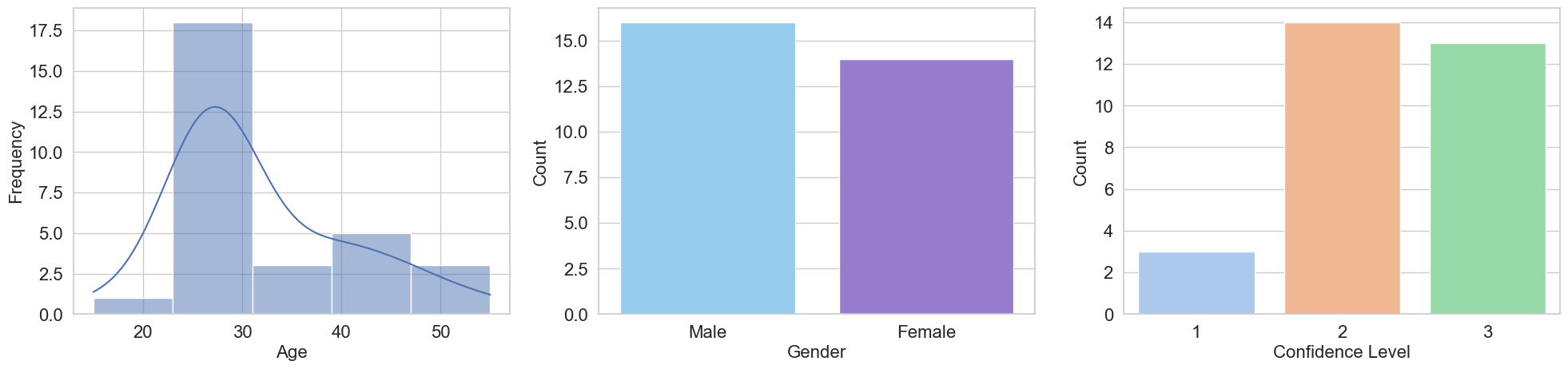}
    \caption{Dataset presentation}
    \label{fig:participants}
\end{figure*}

Please refer to \autoref{fig:participants}.

\section{User Study Analysis Details}
We examine the following hypotheses:
\begin{itemize}
    \item \hypertarget{hp:2}{\textbf{H1}}: The Emotion ON condition will elicit higher levels of engagement and lower levels of boredom in students’ \emph{textual responses}, compared to the Emotion OFF condition.

    \item \hypertarget{hp:3}{\textbf{H2}}: The Emotion ON condition will be associated with a higher frequency of positive \emph{facial expressions} (e.g., happy) and a lower frequency of negative expressions (e.g., sad, angry, fearful etc.) during the tutoring session.

    \item \hypertarget{hp:3}{\textbf{H3}}: The system’s emotion predictions will match user reports with moderate to high agreement.

\end{itemize}

\subsection{Emotion Dynamics in Textual Responses} 

\begin{figure*}[h!]\label{Emotion_Dynamics}
    \centering
    \includegraphics[scale=0.22]{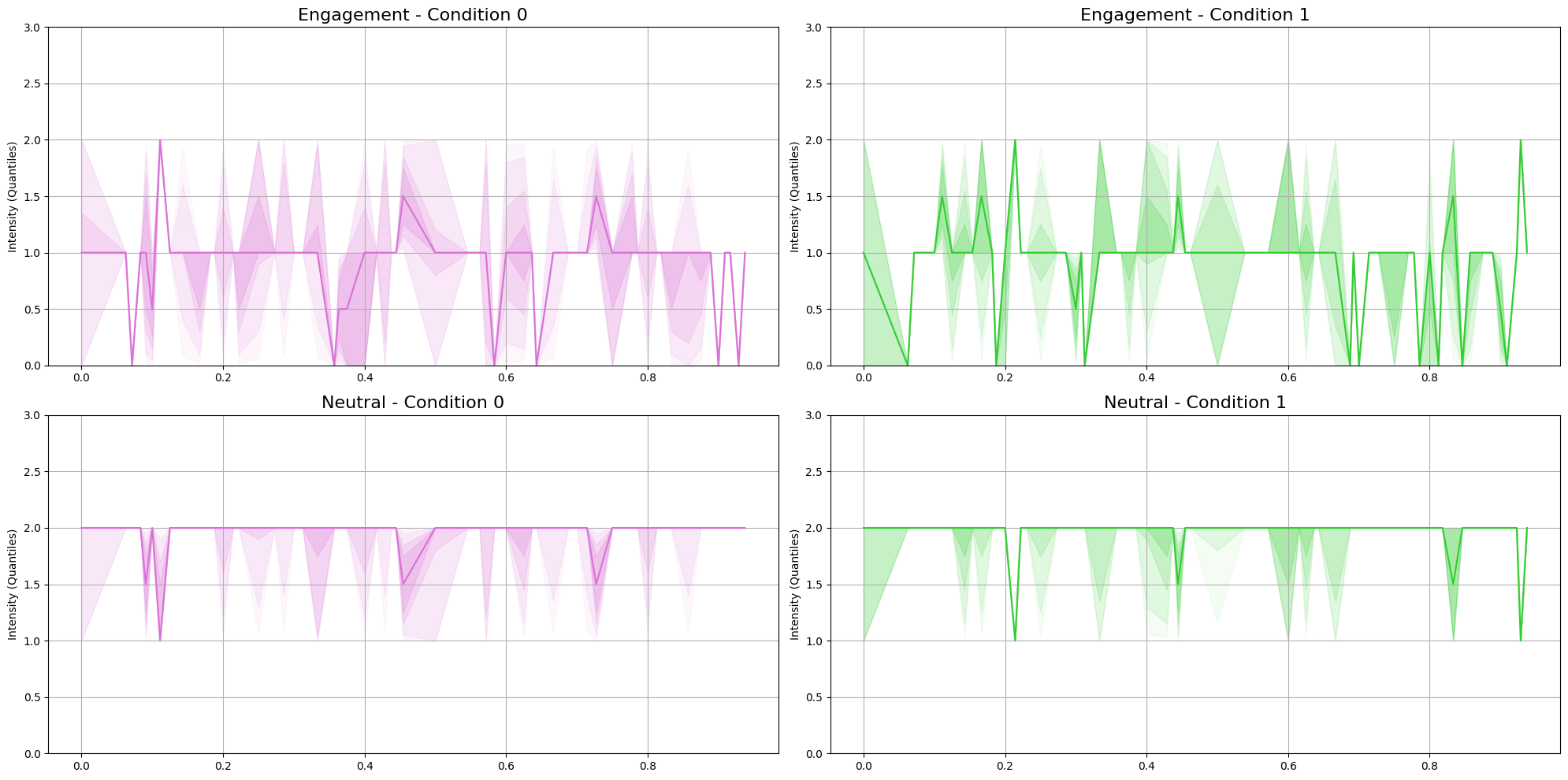}
    \caption{quantile trajectories of textual emotion intensities within interaction sessions.}
    \label{fig:text_quantiles}
\end{figure*}

\begin{figure*}[h!]
    \centering
    \includegraphics[scale=0.24]{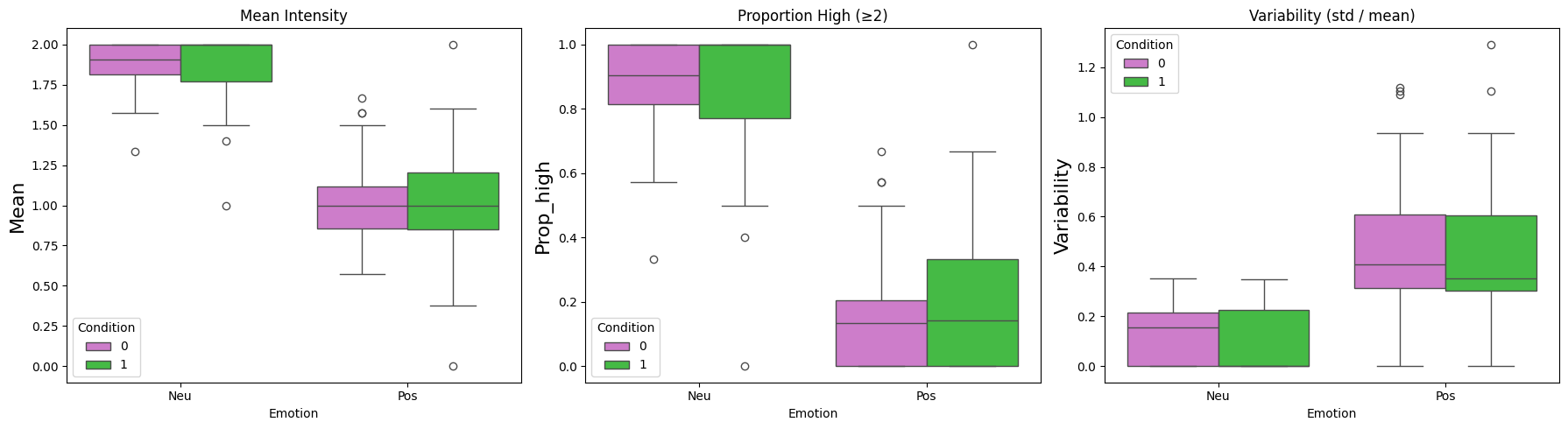}
    \caption{Text emotions statistical properties comparison by condition}
    \label{fig:text_properties}
\end{figure*}

We analyzed 460 student utterances (235 and 225 in the Emotion OFF and ON conditions, respectively), reviewed by the participants themselves for the veracity of the machine-assigned labels. Each utterance was labeled with an intensity score from 0 (not present) to 2 (strong) for each of three emotional categories: \emph{engagement}, \emph{boredom}, and \emph{neutral}.


Further statistical tests are required to confirm the significance of these differences.
Negative emotion is excluded from these as it is never detected by the system in either condition.

\autoref{fig:text_quantiles} displays the temporal quantile plots for engagement and neutral predictions across the two conditions. For each emotion, we report the median intensity (solid line) and quantile bands (shaded areas) over interaction steps.
\begin{itemize}
    \item{Engagement.} Both conditions --- Emotion OFF and ON --- show fluctuating median engagement levels around intensity 1 throughout the session. The Emotion ON condition (upper right) presents slightly more late-session variability, including peaks in the upper quantiles near the final third of the interaction. This may reflect individual differences in how students responded to emotionally adaptive tutoring strategies. The Emotion OFF condition (upper left) shows more symmetric variability distributed throughout the session, without any marked upward trend.
    
    \item{Neutrality.} In both conditions, neutrality maintains high median intensity across the whole session. The Emotion ON one shows slightly more fluctuation in lower quantiles during mid-session relative steps, but the overall pattern suggests a dominant neutral tone throughout, with low between-participant variation.
    
    \item{Boredom.} Predictions were consistently scored as 0 across all utterances and conditions, and are thus omitted from the plot.
\end{itemize}

To complement the trajectory analysis, we ran Wilcoxon signed-rank tests comparing the two conditions (Emotion OFF vs. ON) across three textual emotion metrics: \emph{mean intensity}, \emph{proportion of high-intensity values} ($\ge2$), and \emph{intra-session variability}. Analyses were conducted separately for \emph{neutral} and \emph{positive} emotional classes (\autoref{fig:text_properties}).
Results revealed no statistical significant differences between conditions across any metric, suggesting that the Emotion ON condition did not produce measurable aggregate changes in emotional expression within student text responses.
This suggests that textual input alone may not provide sufficiently informative signals to accurately detect participants' emotions, and that integrating additional multimodal inputs could represent a valuable resource for future improvements.

\subsection{Face Emotion Details}

\begin{figure*}[h!]
    \centering
    \includegraphics[scale=0.22]{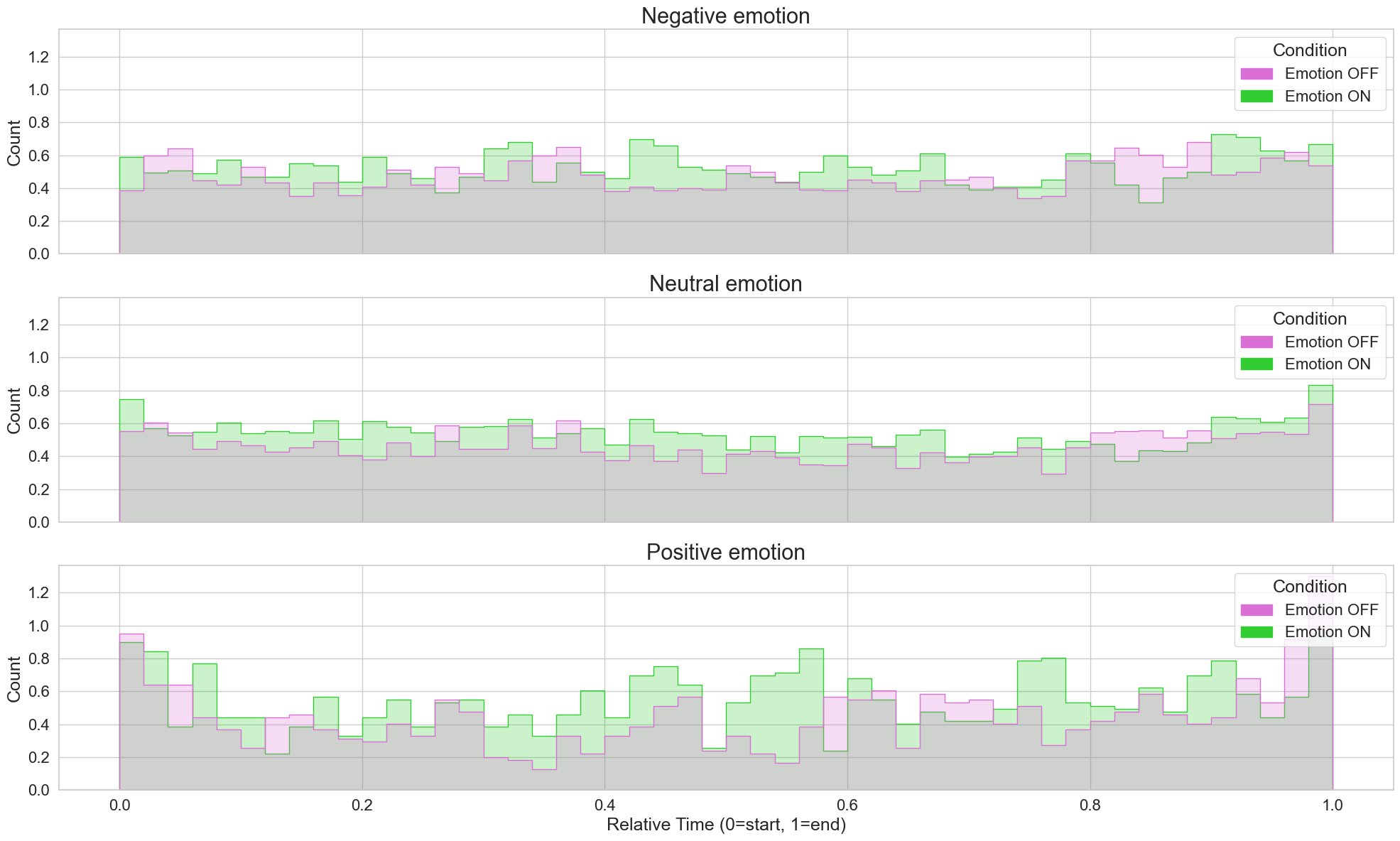}
    \caption{Histogram of facial emotion frequency over normalized session time. Each row shows one emotion category. Color-coded by condition (Emotion OFF = violet, Emotion ON = green).}
    \label{fig:facial_histogram}
\end{figure*}

We analyzed the temporal distribution of students’ facial expressions changes during the tutoring sessions.
Each emotion \emph{positive}, \emph{neutral}, and \emph{negative}, is treated as a discrete categorical value, ignoring the confidence scores.

Given the session lengths and user-specific expressivity, raw emotion timestamps were first converted into discrete time intervals $\Delta' t_{ij}^{(u)}$, defined as the elapsed time between two consecutive detections $t_i$ and $t_j$ within a session for participant user $u$. 
Each $\Delta' t_{ij}^{(u)}$ was assigned to the second timestamp in the pair $t_j$, reflecting the time at which the corresponding emotion was detected.
All timestamps were cumulated and then normalized by the total session duration $\Delta t_\text{max}^{(u)}$ for each users:

\[
\Delta t_{ij,\text{(rel)}}^{(u)} = \frac{\Delta t_{ij}^{(u)}}{\sum_{ij} \Delta t_{ij}^{(u)}} = \frac{\Delta t_{ij}^{(u)}}{\Delta t_\text{max}^{(u)}}\, ,
\]

where $\Delta t_{ij}^{(u)}$ is the cumulative time from the start of the session up to the $j$-th detection. The result is a temporally standardized view of facial emotion dynamics across all participants.
This normalization enabled comparison of emotion frequencies across participants on a common temporal axis ranging from 0 (session start) to 1 (session end).

\begin{figure*}[h!]
    \centering
    \includegraphics[scale=0.22]{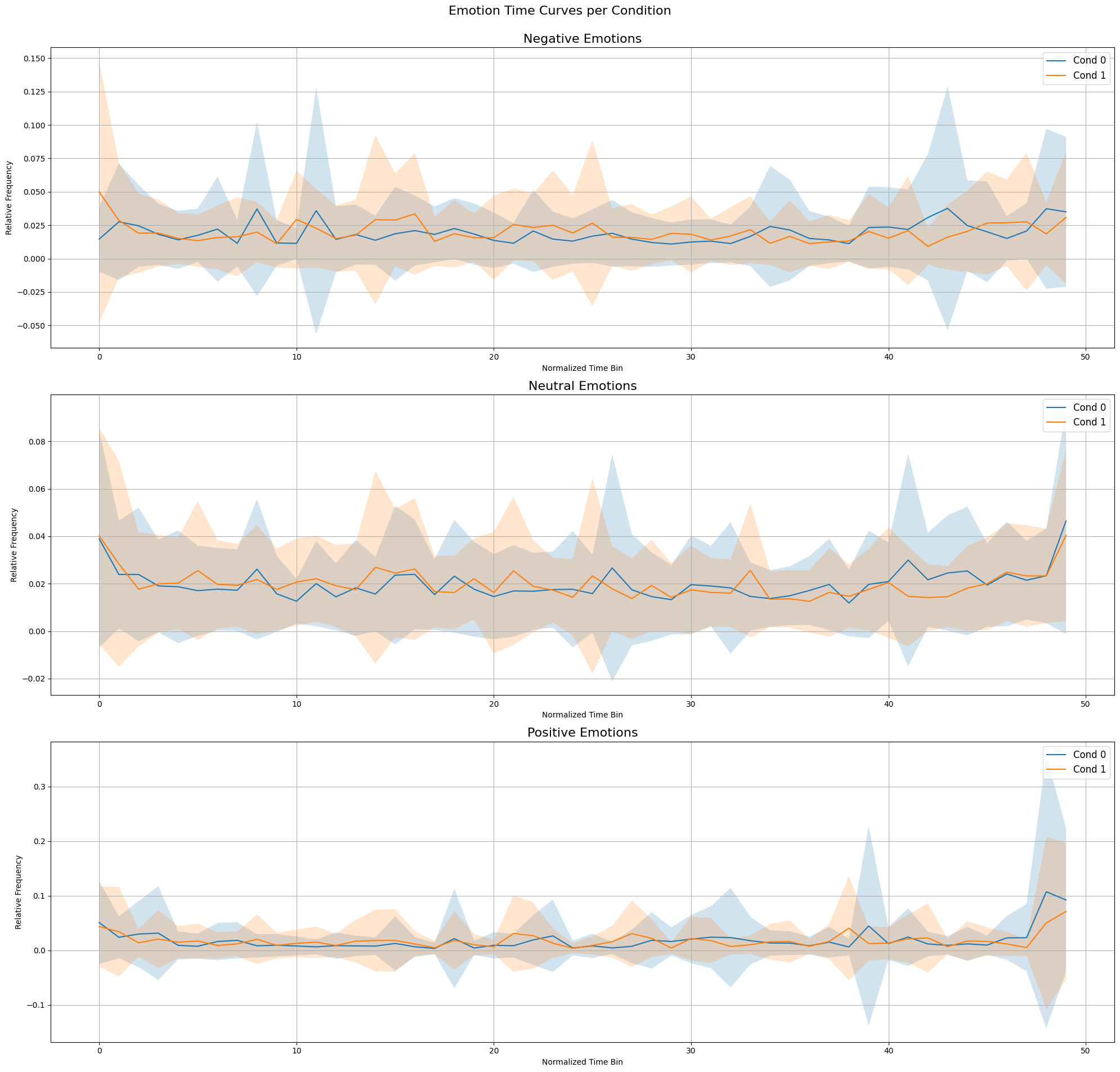}
    \caption{Average temporal curves of facial emotion frequency per participant, with standard deviation}
    \label{fig:facial_curves}
\end{figure*}

\autoref{fig:facial_histogram} shows the relative density of detected emotions aggregated across participants and grouped into temporal bins, while \autoref{fig:facial_curves} show the resulting mean time series with shaded areas representing one standard deviation, separately for Emotion OFF (blue) and Emotion ON (orange) conditions.
These curves capture the general trends in emotion distribution over time and the degree of agreement among participants. A wide standard deviation band indicates higher variability—i.e., some users expressed the emotion frequently in that bin, others not at all. 
Overall, both conditions show overlapping trends, with slight increases in positive emotions at the end of the session under Emotion ON. However, variability remains high across participants, especially for positive expressions, limiting the strength of conclusions.

To statistically test for distributional differences between conditions, we computed two shape-based metrics per user and emotion class:
\begin{itemize}
    \item \textbf{Centroid}: the weighted temporal center of the emotion distribution;
    \item \textbf{Skewness}: the temporal distribution asymmetry (positive = delayed, negative = early).
\end{itemize}

Results from Wilcoxon signed-rank tests revealed no significant differences in centroid position or skewness between Emotion ON and OFF across all emotion classes ($p > 0.05$). 
Descriptively, the centroid for positive emotions was slightly earlier in Emotion ON (mean = 26.2) than in OFF (mean = 27.9), and skewness slightly lower (2.42 vs. 2.79), suggesting a minor temporal anticipation of positive expressions.

\begin{figure*}[h!]
    \centering
    \includegraphics[scale=0.4]{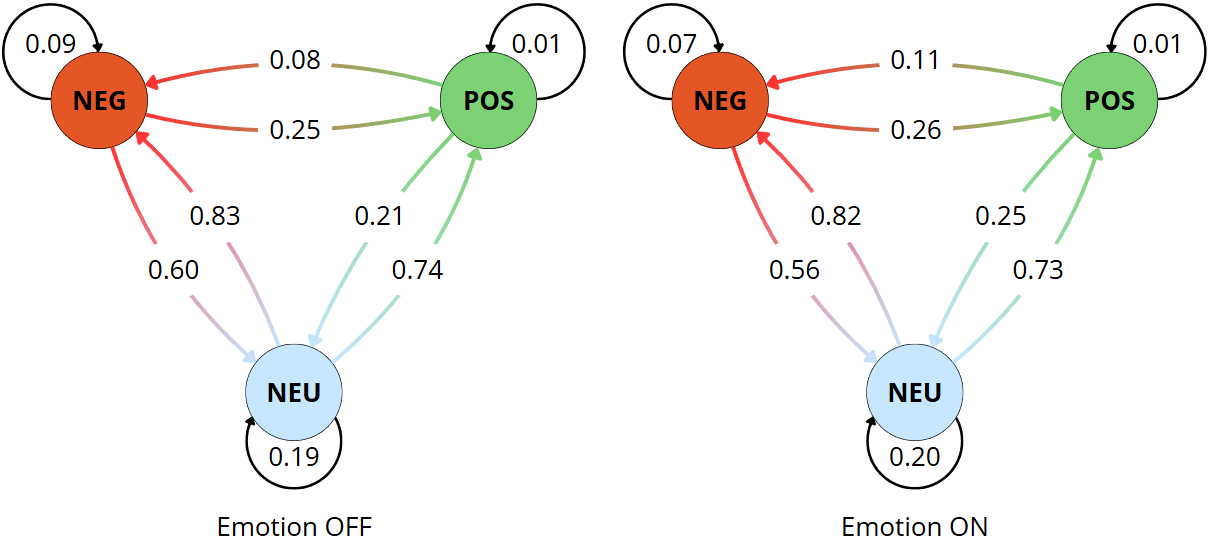}
    \caption{Transition graphs for facial expression emotion by condition.}
    \label{fig:graph_transitions}
\end{figure*}

To investigate the structural patterns of emotional flow during the sessions, we analyzed the transitions between facial emotion classes. Each session was modeled as a sequence of discrete emotional states sorted by cumulative timestamp.
For each participant and condition, we computed a transition matrix capturing the frequencies of all possible emotion-to-emotion transitions (e.g., \texttt{Pos$\rightarrow$Neu}).
Transition counts were normalized by the total number of transitions per user, and then averaged across participants to obtain group-level transition probabilities. The resulting transition graphs are shown in \autoref{fig:graph_transitions}, separately for the Emotion OFF and ON conditions.
Each node represents an emotional state, with node size indicating average state occupancy. Directed edges indicate transitions between states, with edge thickness and labels proportional to transition probability.

We further analyzed the average duration of each facial emotion under both conditions. For every emotion occurrence, duration was computed as the time elapsed until the next detection ($\Delta t = t_{i+1} - t_i$), and assigned to the first timestamp $t_i$, under the assumption that the emotion persisted until a new one was detected.

\subsection{System's ability to detect emotions}
To test last hypothesis, we evaluated the system's predictions against gold-standard annotations and computed standard classification metrics. As a final step, participants in the Emotion ON condition reviewed their own utterances at the end of the experiment and adjusted the tutor’s emotion labels when necessary. This procedure provided a reliable basis for assessing the system’s emotion recognition performance.
We compared the system-generated emotion labels with the participant-corrected annotations and computed standard classification metrics. The system achieved an overall accuracy of $60$\%, with particularly high recall for the negative class ($98$\%), but lower performance for neutral and positive classes, especially in terms of recall. This suggests that while the system is effective at detecting negative emotions, it struggles to consistently identify neutral or positive affect, which are more subtle or context-dependent.
See table \autoref{tab:emotion_classification}.

\end{document}